\documentclass[11pt]{article}
\usepackage{CJKutf8}
\usepackage{CJKspace}  

\usepackage{ACL2023}

\usepackage{times}
\usepackage{verbatim}
\usepackage{latexsym}
\usepackage{algorithm}
\usepackage{algpseudocode}
\usepackage{enumerate}
\usepackage{fancyvrb}
\usepackage{multicol}
\usepackage{listings}
\usepackage{cite}
\usepackage{enumitem}
\usepackage{graphicx}
\usepackage{caption}
\usepackage{subcaption}
\usepackage{alltt}
\usepackage{cleveref}
\usepackage{booktabs}
\usepackage{svg}
\usepackage[]{mdframed}

\usepackage[T1]{fontenc}

\usepackage[utf8]{inputenc}

\AtBeginDocument{%
  \begin{CJK*}{UTF8}{gbsn}%
}
\AtEndDocument{%
  \end{CJK*}%
}
\newcommand{\zh}[1]{#1}

\usepackage{microtype}

\usepackage{inconsolata}
\usepackage{etoolbox}
\makeatletter
\patchcmd{\@verbatim}
  {\verbatim@font}
  {\verbatim@font\scriptsize}
  {}{}
\makeatother

\usepackage{array,etoolbox}
\preto\tabular{\setcounter{magicrownumbers}{0}}
\newcounter{magicrownumbers}
\def\rownumber{}

\title{Optimal Corpus Aware Training for Neural Machine Translation}

\author{
  Yi-Hsiu Liao \\
  \texttt{yihsiu\_liao@apple.com}
  \\\And
  Cheng Shen \\
  \texttt{cshen2@apple.com} \\
  \\
   Apple 
  \\\And
  Brenda (Zixiaofan) Yang \\
  \texttt{zixiaofan\_yang@apple.com}
}

\begin{document}

\maketitle

\begin{abstract}
Corpus Aware Training (CAT) leverages valuable corpus metadata during training by injecting corpus information into each training example, and has been found effective in the literature, commonly known as the "tagging" approach. Models trained with CAT inherently learn the quality, domain \& nuance between corpora directly from data, and can easily switch to different inference behavior. To achieve the best evaluation, CAT models pre-define a group of high quality data before training starts which can be error-prone and inefficient. In this work, we propose Optimal Corpus Aware Training (OCAT), which fine-tunes a CAT pre-trained model by freezing most of the model parameters and only tuning small set of corpus-related parameters. We show that OCAT is lightweight, resilient to overfitting, and effective in boosting model accuracy. We use WMT’23 English to Chinese and English to German translation tasks as our test ground and show +3.6 and +1.8 chrF improvement, respectively, over vanilla training. Furthermore, our approach is on-par or slightly better than other state-of-the-art fine-tuning techniques while being less sensitive to hyperparameter settings.



\end{abstract}

\section{Introduction}

\begin{figure*}
\centering
\begin{subfigure}{.5\textwidth}
  \centering
  \includegraphics[width=.95\linewidth]{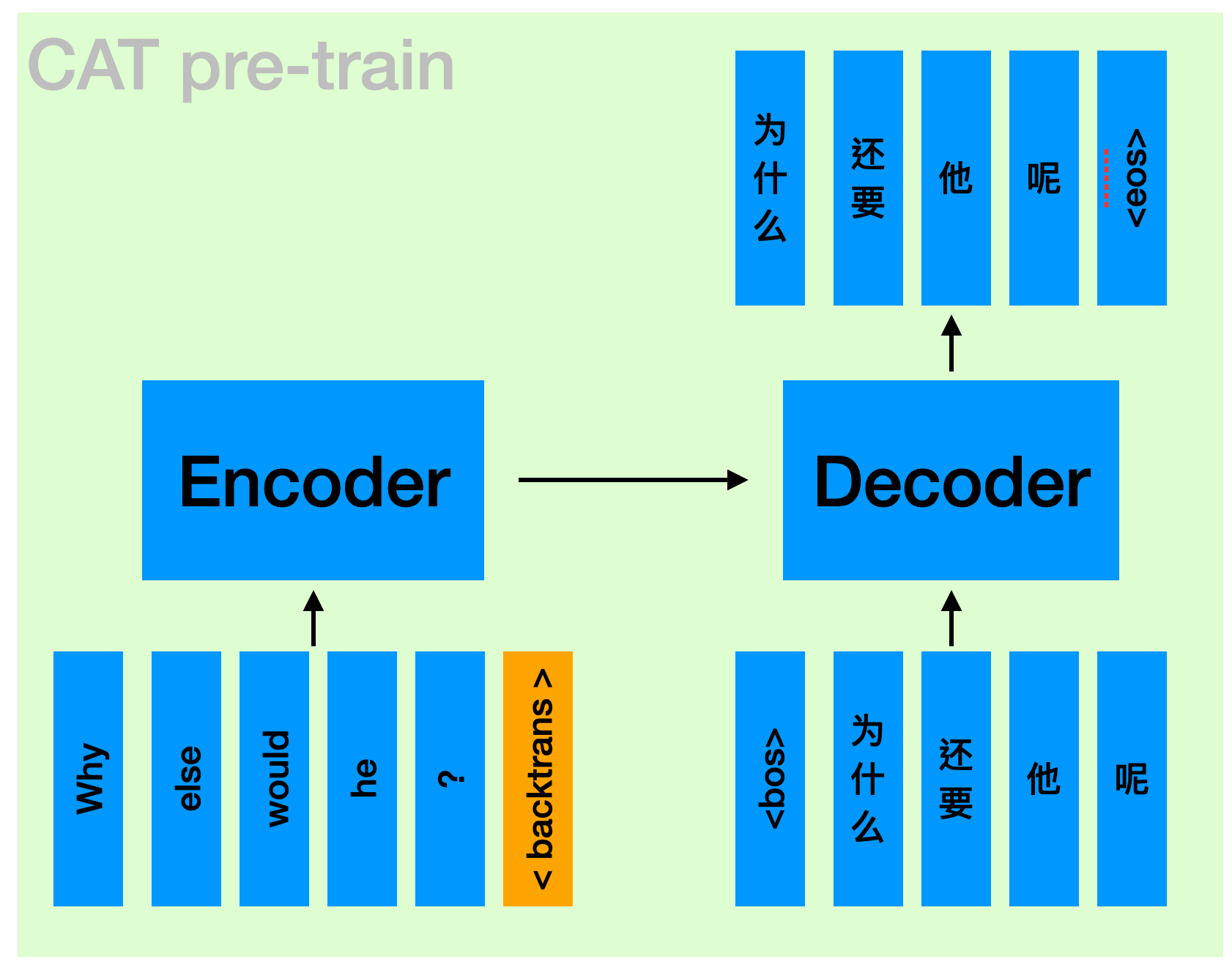}
\end{subfigure}%
\begin{subfigure}{.5\textwidth}
  \centering
  \includegraphics[width=.95\linewidth]{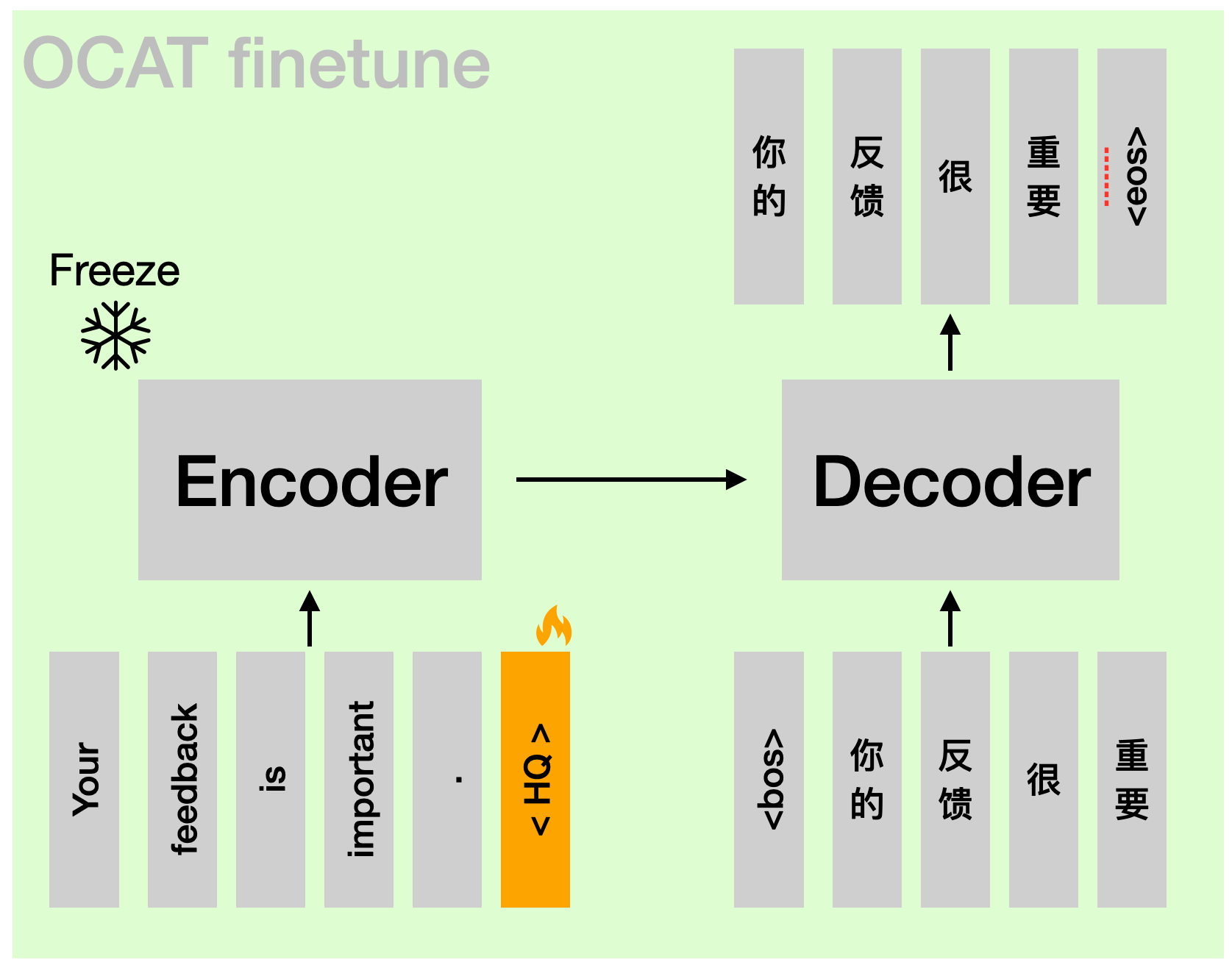}
\end{subfigure}
\caption{The main difference to typical training scheme is highlighted in orange where the CAT inserts the corpus tags per example and OCAT freezes all parameters except one corpus tag token embedding. The <HQ> token means high quality corpus tag that will be selected for inference.}
\label{fig:ocat_diagram}
\end{figure*}


In a typical Machine Learning (ML) training recipe, practitioners usually mix all available datasets together to build a large training dataset, including acquired, licensed, crawled, and manually annotated datasets. The domain, quality and quantity of collected corpora are substantially different, and blindly training the whole mixture loses valuable corpus information, leading to a sub-optimal result.

There are several works \citep{kobus2016domain,tars2018multi,caswell2019tagged,sennrich2016controlling,costa2022no} which utilize valuable corpus information and enable the model to learn the nuances between different corpora. In this paper, we refer to this generic approach as Corpus Aware Training (CAT), which focuses on the differences in corpora distribution. 
By inserting corpus information into each training example as tags, the CAT model becomes aware of the origin of each example. Each corpus tag denotes a unique mode, in which model inference runs. Therefore, CAT models are capable of switching between different prediction modes during inference. Our results verify that mode-switching significantly influences evaluation metrics. 

Given the effectiveness of CAT, we face a natural practical question -- how do we decide which inference mode to use for model deployment? Or similarly, what is the optimal inference mode that can best generalize to testing domains?

We address the above questions with a generic training scheme that boosts CAT models by leveraging small, clean, high-quality in-domain data -- optimal CAT (OCAT). Drawing inspiration from \citet{li2021prefix}, we only fine-tune corpus-related parameters and freeze all other model parameters, illustrated in \Cref{fig:ocat_diagram}. We show that OCAT is lightweight and only needs less than a single A100 GPU-hour to converge. Yet, the proposed approach seems to avoid overfitting due to the limited trainable parameter and helps to improve a CAT pretrained model.

Experiments have been conducted on two common translation tasks (WMT’23 English to Chinese and English to German) employing the Transformer architecture  \citep{vaswani2017attention}. Our experiments show +3.6 and +1.8 chrF improvement, respectively, compared to training without CAT.

\section{Related Work}
There are many works that uses the tagging approach but does not focus on fine-tuning a CAT pre-trained model. \citet{sennrich2016controlling} introduces additional token per training example to control politeness level in translation, \citet{johnson2017google} proposes to use a similar tagging approach to control translation direction in multilingual neural machine translation. \citet{kuczmarski2018gender} also use per example tags to control gender in translation. \citet{kobus2016domain} insert domain information by either additional tag or feature embedding to build translation models. \citet{tars2018multi} add domain information to each training example and open a new direction to classify domain and explore both supervised and unsupervised approaches to cluster domains. \citet{saunders2022domain} summarizes most popular approach to domain adaptation and multi-domain adaption with tags.

\citet{caswell2019tagged} tagged back translation data differently from the original parallel data, and shows significant improvement. \citet{costa2022no} extended the above tagging schemes to separate back translation data from neural machine translation or statistical machine translation. \citet{freitag-etal-2022-natural} improves translation quality by tagging training examples with either natural or translationese. \citet{tomani2024quality} trained quality-aware prompting models which tagged each training example by its quality score from an external QE model. \citet{saxena2019data} pushes tagging approach to a finer sample level, that labels each training sample with separate parameters, and does not cluster training samples.

Our proposed OCAT draws inspiration from \citet{li2021prefix,lester2021power} that fine-tunes limited parameters from pretrained models to guide LLM \citep{radford2019language,devlin2018bert} with small amounts of fine-tune data. OCAT also freezes most parameters and tunes limited parameters but only applies to the CAT pretrained model. OCAT resolves the inference mode selection problem and generalize the CAT pretrain model towards multiple test domains.

\section{Method}
\subsection{Corpus Aware Training}
\label{sec:CAT}

CAT inserts corpus information to each training input example so that given any input, the model is aware where this example comes from. We use neural machine translation as our main experiment setup, and follow the tagging approach in \citep{kobus2016domain} to append a corpus name tag to the end of each source sentence, eg. <OPUS-v1.0>, <backtrans>, <paracrawl-v9>, etc.




During inference, we pre-select a decoding corpus tag (eg. <backtrans>) that is used in training. We name the tag used during inference as \textit{inference tag} in the remaining of this paper. In \Cref{sec:cat_exp}, we show that the same CAT model decodes with different inference tag can differ up to 8.5 chrF points. This shows CAT is effective in capturing distribution difference and a wrong choice of inference tag could lead to severe accuracy drop.

To achieve the optimal evaluation score on target testset, we enumerate all training corpus tags and select the best inference tag by dev set score. However, inference tag selection is not trivial if there are multiple inference tags that share similar high scores or if there are multiple domains we want to target for deployment. One naive solution to optimize this is re-training the model from scratch with each sample from selected corpora tagged with <HQ> (high quality tag). However, complete re-training is costly and hard to scale.

\subsection{Optimal Corpus Aware Training}
While CAT is effective in switching inference mode, it doesn't provide a straightforward way to select the best inference tag to fit multiple test domains. We introduce OCAT below to mitigate the inference tag selection problem.

Inspired by \citet{li2021prefix}, we show that freezing all the parameters except the corpus related parameters (tag embeddings) is efficient and effective to optimize CAT pretrained model. We name the proposed approach Optimal Corpus Aware Training that helps to find the optimal inference tags with minimal resource requirement. The number of trainable parameters for OCAT is equal to the dimension of a single token embedding. For example, if we use transformer base \citep{vaswani2017attention}, there are only 512 parameters to be fine-tuned (out of 69M parameters). Due to limited trainable parameters, the model is not capable of overfitting even if fine-tune dataset is small.

OCAT is lightweight. The transformer base model converges in less than one hour if fine-tuned on a single A100 GPU. OCAT is also resilient to training instability and fine-tune hyper-parameter. Although we only tuned inference tag embedding, OCAT is effective in improving the model prediction because CAT models are sensitive to the choice of inference tag as mentioned in \Cref{sec:CAT}, and tuning tag embedding steers model behavior towards the desired direction.

In practice, we combine several high quality training datasets along with validation sets as OCAT fine-tuning dataset to generalize to multiple test domains. In the WMT23 en-zh setup, however, we found that even the best training dataset does not bring as much improvement as validation dataset. In this case, dataset combination step is optional and using validation alone as OCAT fine-tune dataset could be sufficient.

We propose a generic ML training process in \Cref{alg:ocat}.

\begin{algorithm}
\caption{OCAT training scheme}\label{alg:ocat}
\begin{algorithmic}[1]

\State Pretrain with Corpus Aware Training
  \begin{enumerate}[label=(\alph*)]
    \item Insert corpus info to training examples.
  \end{enumerate}
\State Build OCAT fine-tune dataset
  \begin{enumerate}[label=(\alph*)]
    \item Evaluate training corpora quality by enumerating corpus tags.
    \item Select a few corpora that achieves highest validation score.
    \item (Optional) Combine selected corpora with validation set as OCAT fine-tune dataset.
  \end{enumerate}
\State fine-tune tag embeddings on selected datasets.

\end{algorithmic}
\end{algorithm}

\section{Experiment}
\subsection{Experimental Setup}

We follow the WMT 23 en-zh setup \citep{tom2023findings}, and leave WMT23 en-de results in the \Cref{sec:en_de}. We use all WMT available parallel data with 20M sentences limit per corpus to avoid being overwhelmed by a single large quantity automatically crawled dataset. We do not use any monolingual data nor other data source outside WMT. We do not use CCMT corpus because of restricted access. There are 6 parallel corpora in en-zh, therefore, we insert 6 customized tag tokens to the dictionary. We combine newstest 22 and flores.dev as fine-tune dataset and use newstest23 as our testset and newstest22 as our validation set, we also include flores as additional testset. We use chrF \citep{popovic2015chrf} from sacrebleu \citep{post-2018-call}\footnote{nrefs:1|case:mixed|eff:no|tok:13a|smooth:exp|version:2.1.0} as our main evaluation metric, and report metricX \citep{juraska-etal-2023-metricx} \footnote{METRICX-metricx-23-qe-large-v2p0} only for reference because the metricX delta can be too small to be significant.

We use \textsc{fairseq} \citep{ott2019fairseq} to train a vanilla transformer base encoder-decoder model with 6 encoder layers, 6 decoder layers. Hidden dimension is 512, with feedforward layer dimension equal to 2048. There are 8 attention heads in all multi-head attention blocks and each head dimension equals to 64. We share the encoder, decoder word embedding and decoder output projection to minimize memory footprint. We train a joint sentencepiece model \citep{kudo2018subword} with vocab size equals to 48k for source and target language combined as our tokenizer. The number of parameters in our translation model is 69M. We train 100k steps with the first 4000 steps linear warmup to maximum learning rate=4e-4 and then decay with inverse square root. We use Adam optimizer \citep{kingma2014adam} with betas = (0.9, 0.98). The batchsize is around 500k tokens. We save checkpoints every 2000 steps and average the final 5 checkpoints as our final model.

\begin{table*}[htb]
\centering
\begin{tabular}{@{\makebox[1.5em][l]{\rownumber\space}}llcccc}\toprule
 Model    & inference tag         & nt23 & nt22 & flores devtest & flores dev
 \gdef\rownumber{\stepcounter{magicrownumbers}\arabic{magicrownumbers}} \\\midrule
 baseline & -                     &  40.2 / 2.9  &  34.4 / 2.0  &  34.4 / 1.9  &  33.4 / 2.0  \\\midrule[0.1pt] 
 + CAT & <news-commentary-v18> &  42.5 / 1.8  &  37.1 / 1.5  &  37.3 / 1.4  &  36.8 / 1.4  \\ 
         & <wikititles-v3>       &  39.2 / 2.2  &  32.9 / 1.9  &  32.0 / 1.5  &  31.2 / 1.6  \\ 
         & <OPUS-v1>             &  38.2 / 2.1  &  33.8 / 1.7  &  34.7 / 1.5  &  34.1 / 1.6  \\ 
         & <wikimatrix-v3>       &  33.9 / 2.6  &  29.5 / 2.2  &  29.5 / 1.8  &  29.2 / 1.7  \\\midrule[0.1pt] 
+  OCAT &                       &  \textbf{44.0} / 1.8  &  \textbf{38.7} / 1.5  &  \textbf{37.9} / 1.4  &  \textbf{37.2} / 1.4  \\ 
         \bottomrule
\end{tabular}
\caption{chrF/metricX score for WMT 23 $En \rightarrow Zh$ trained w/ \& w/o CAT and follow OCAT fine-tuning with newstest22 + flores.dev. Here, nt is newstest. We use \textbf{bold} font to mark the top-1 chrF score that is statistical significant better than the runner-up with p-value <= 5\%.}
\label{table:cat}
\end{table*}

\subsection{Results and Analysis}
\subsubsection{Model prediction is sensitive to inference tag}
\label{sec:cat_exp}
We train a baseline model without CAT and a model with CAT for English to Chinese and English to German and report the numbers in \Cref{table:cat}. Comparing rows 2 to 5, decoding with different inference tag impacts model prediction by up to 8.6 chrF points / 0.8 metricX and can be used as a reliable way to switch between different decoding mode. Among all inference tags, decoding with <news-commentary-v18> shows the best chrF score which also implies the news-commentary-v18 training data is the highest quality dataset. We report a few inference tags in row 3 and 4, and observe that chrF score fluctuates more compared with metricX and some minor difference might be only captured by chrF.

\begin{table*}[htb] 
\centering
\def\rownumber{}
\begin{tabular}{@{\makebox[1.5em][l]{\rownumber\space}}lclcccc}\toprule
 Model   & trainable params & fine-tune data           & nt23 & nt22 & flores devtest & flores dev
 \gdef\rownumber{\stepcounter{magicrownumbers}\arabic{magicrownumbers}} \\\midrule
 w/  CAT & -                & -                        &  42.5 / 1.8  &  37.1 / 1.5  &  37.3 / 1.4  &  36.8 / 1.4  \\ 
+   OCAT & 512              & nc + nt22 + flores.dev   &  42.7 / 1.8  &  37.3 / 1.5  &  37.3 / 1.4  &  37.0 / 1.4  \\ 
+   OCAT & 512              & nt22 + flores.dev        &  44.0 / 1.8  &  38.7 / 1.5  &  37.9 / 1.4  &  37.2 / 1.4  \\ 
+   OCAT & 512              & flores.dev               &  43.2 / 1.9  &  38.0 / 1.5  &  38.0 / 1.5  &  37.7 / 1.4  \\ 
+   OCAT & 512              & nt22                     &  44.0 / 1.8  &  39.0 / 1.5  &  37.5 / 1.4  &  37.0 / 1.4  \\ 
\midrule
 w/o CAT  & -               & -                        &  40.2 / 2.9  &  34.4 / 2.0  &  34.4 / 1.9  &  33.4 / 2.0  \\ 
+ full FT & 69M             &   nt22                   &  33.9 / 2.2  &  \textbf{99.9} / 1.2  &  30.2 / 1.8  &  29.9 / 1.8  \\ 
+ adapter & 246k            &   nt22                   &  43.8 / 2.0  &  41.9 / 1.5  &  37.8 / 1.4  &  36.0 / 1.4  \\ 
+ LoRA    & 1.2M            &   nt22                   &  42.5 / 1.9  &  51.9 / 1.4  &  37.3 / 1.4  &  36.1 / 1.4  \\ 
         \bottomrule
\end{tabular}
\caption{chrF/metricX scores for WMT 23 $En \rightarrow Zh$. We choose <news-commentary-v18> as the inference tag for CAT and apply OCAT with different fine-tune datasets. Here, nc is news-commentary-v18, nt22 is newstest22. We also compare against training without CAT and fine-tune with full-weight, adapter and LoRA. Among the 3 popular techniques, adapter is on par with OCAT for in-domain (nt22 \& nt23).}
\label{table:ocat}
\end{table*}

\subsubsection{Apply OCAT with the proposed training scheme}
If we want to deploy a model that performs best across multiple testsets, e.g.~newstest08-23 and flores, we can leverage OCAT to tune corpus tag embedding so it fits multiple domains better. We select a few corpora corresponding to top performing corpus tag for newstest23 and flores, and combine selected training corpora with validation set as OCAT fine-tuning dataset. In practice, if we have many high quality training data, the proposed scheme is more effective. In WMT23 en-zh, however, the highest quality training data, news-commentary-v18, lags significantly behind validation set quality, see \Cref{table:ocat} row 2 and row 3. If we only tune corpus tag with validation set (newstest22 and flores.dev), we see +3.6 chrF improvement in newstest23 and +3.6 chrF improvement in flores.devtest over baseline model w/o CAT as shown in \Cref{table:ocat} row 3 and 6.

\subsubsection{OCAT is resilient to overfitting}

In general, we should avoid using validation data as our training/fine-tune dataset to prevent overfitting. It is unconventional to directly train on validation set while making model decision still with the same validation set. However, OCAT fine-tunes a limited number of parameters, and only changes the corpus tag embedding. We argue that OCAT is resilient to overfitting and would not be constraint by the above rule.
We apply OCAT with only newstest22 or flores.dev as fine-tune dataset and report model performance in \Cref{table:ocat}. Tuning on validation set does not degrade other testset chrF and only boosts the chrF score of tuning set by +4.6, +4.3 chrF for newstest22 and flores.dev respectively in row 4, 5, 6.

\begin{figure*}[h]
\centering
\begin{subfigure}{.5\textwidth}
  \centering
  \includegraphics[width=.95\linewidth]{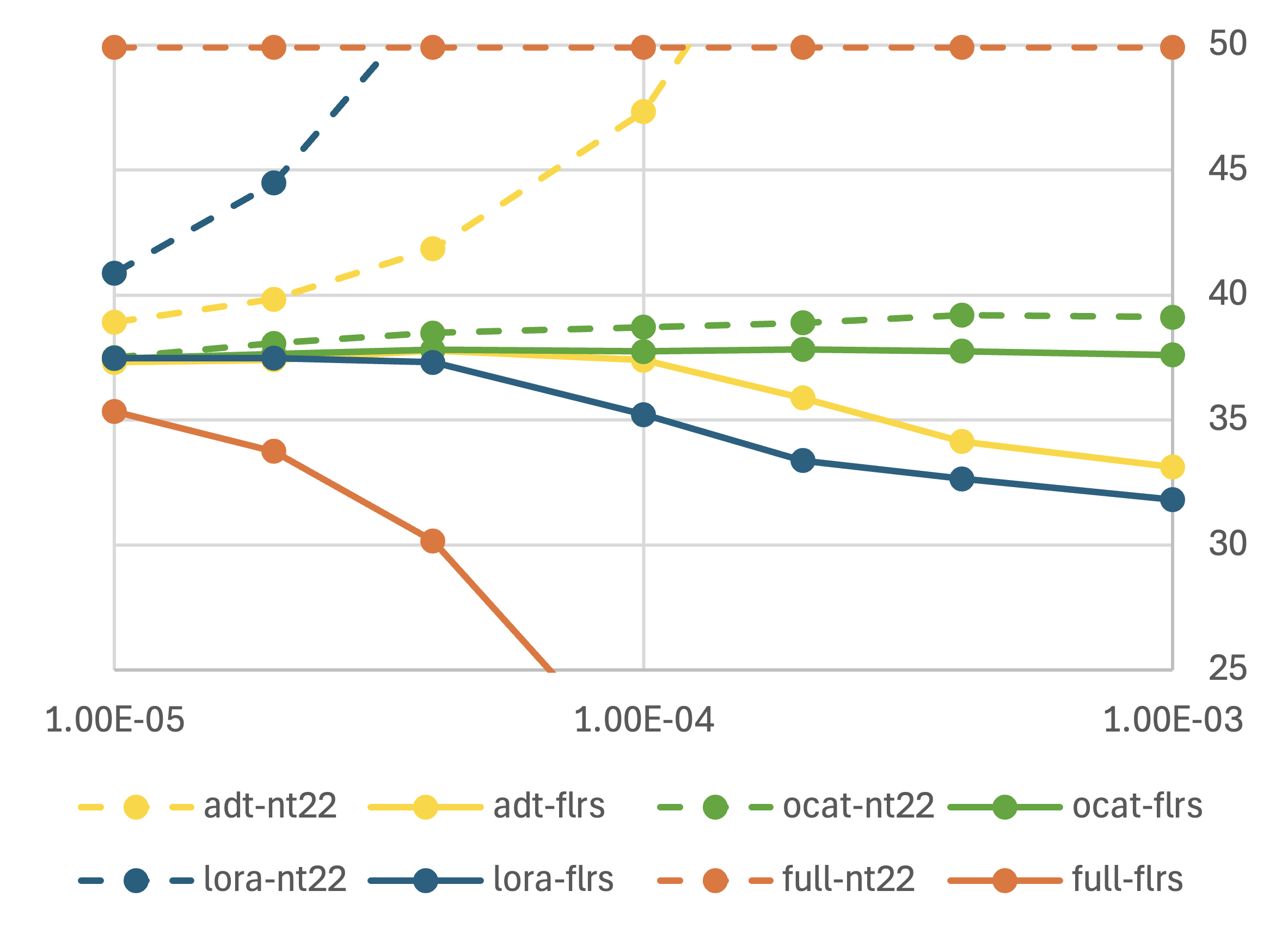}
  \caption{fine-tune learning rate vs chrF score}
\end{subfigure}%
\begin{subfigure}{.5\textwidth}
  \centering
  \includegraphics[width=.95\linewidth]{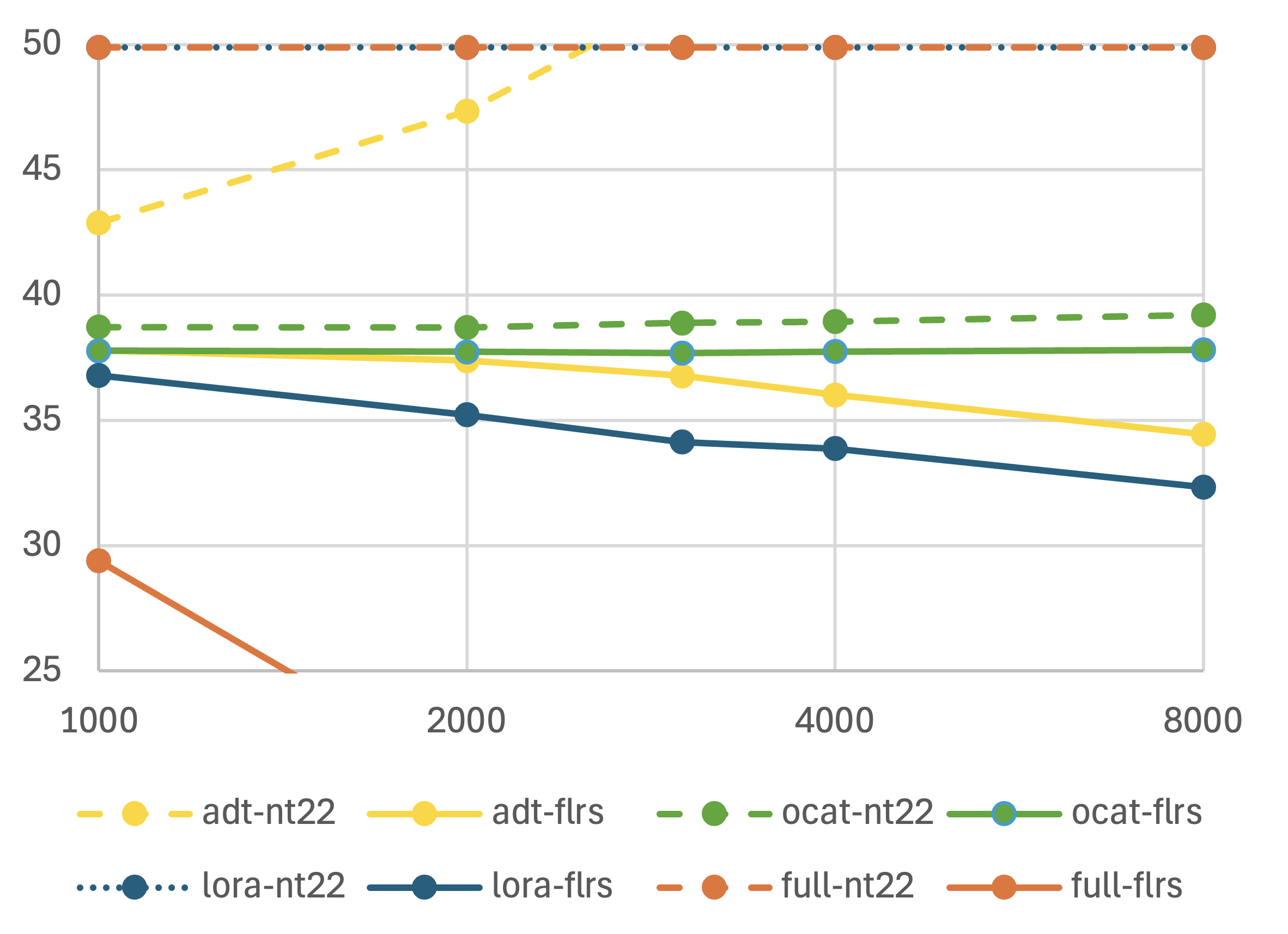}
  \caption{fine-tune training steps vs chrF score}
\end{subfigure}
\caption{We change hyper-parameters in fine-tune stage, group different fine-tune approach by color and use solid lines to mark out-of-domain testset and dash lines are training data performance. OCAT remains stable (flat) for a wide range of hyper-parameters and does not overfit to training data. Full fine-tune overfits easily. Both adapter and LoRA suffers from overfitting with higher learning rate and larger number of steps.}
\label{fig:stability}
\end{figure*}

\subsubsection{Compare OCAT against other fine-tuning techniques}
We compare OCAT with other domain adaptation techniques applied to non-CAT pretrained models including full-weight, adapter \citep{houlsby2019parameter}, LoRA \citep{hu2022lora} fine-tuning which are reported in \Cref{table:ocat} row 6-9. We can see that full-weight fine-tuning overfits to tuning set (newstest22) and degrades significantly on newstest23 and flores. Adapter and LoRA showed much better generalization ability but still lags behind OCAT in out-of-domain testsets (flores). Another interesting finding is that, there are only 2k sentences in newstest22, that adds another challenge to stably fine-tune a pretrained model. OCAT, on the other hand, does not have this constraint and stays resilient to a wide range of hyper-parameter setups, details in \Cref{fig:stability}.

\begin{figure}[h]
\includegraphics[scale=0.2]{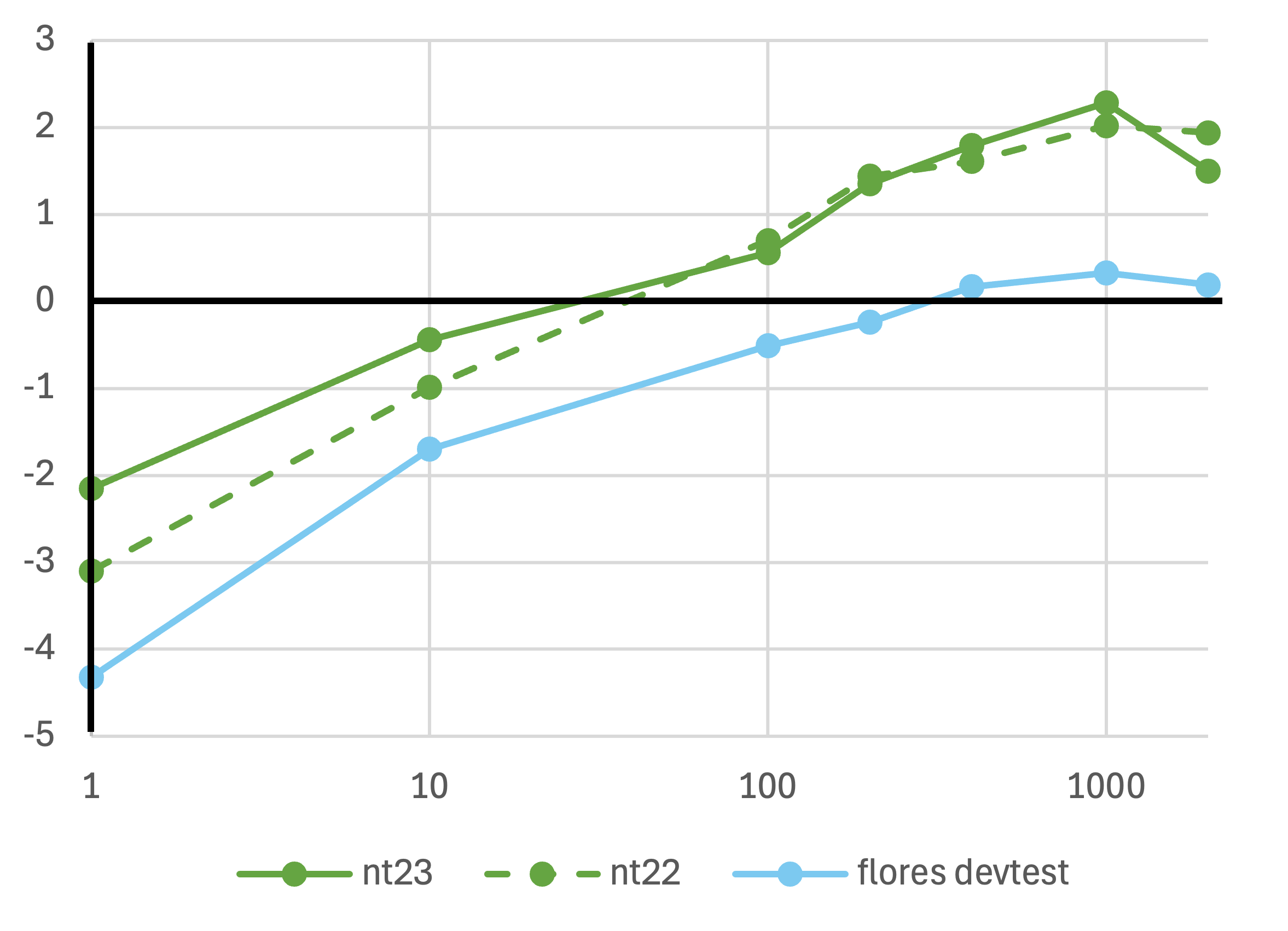}
\caption{WMT23 en-zh, OCAT chrF improvements trained with n sentences sampled from newstest22.}
\label{fig:fine-tune_size}
\end{figure}

\subsubsection{The minimum size of OCAT fine-tune dataset}
We would like to push the limit of OCAT and find out the minimum size of OCAT fine-tune dataset. In \Cref{fig:fine-tune_size}, we fine-tune OCAT corpus tag embedding by as little as 1 sentence as fine-tune dataset up to 2k sentences, the full newstest22 set. To our surprise, tuning with 1 sentence does not degrade the model too much on in-domain testsets (newstest23, newstest22) and starts to show improvements with as few as 100 sentences. For out-of-domain testset (flores) that was not included in WMT23 training and evaluation, OCAT requires at least 400 sentences to avoid degradation and see minor improvements. This validates our proposed training scheme that even though OCAT is not data-hungry and converges easily, in practice, if we include more high quality data in the OCAT fine-tune stage, the tuned corpus tag embedding is more capable of generalizing to more test domains.

\begin{table*}[htb]
\centering
\def\rownumber{}
\begin{tabular}{@{\makebox[1.5em][l]{\rownumber\space}}lcccc}\toprule
 Decode tag            & nt23 & nt22 & flores devtest & flores dev
 \gdef\rownumber{\stepcounter{magicrownumbers}\arabic{magicrownumbers}} \\\midrule
<news-commentary-v18>  &  \textbf{42.3} / 1.8  &  \textbf{36.7} / 1.5  &  \textbf{37.1} / 1.4  &  \textbf{36.7} / 1.4  \\ 
<ParaCrawl-baunat>     &  39.7 / 1.9  &  34.5 / 1.6  &  34.1 / 1.5  &  33.2 / 1.5  \\ 
<ParaCrawl-php>        &  34.9 / 3.7  &  31.0 / 3.0  &  31.1 / 2.6  &  30.9 / 2.5  \\ 
<ParaCrawl-5zixi>      &  10.9 / 12.8  &  13.6 / 11.6  &  12.8 / 14.4  &  13.0 / 14.1  \\ 
\bottomrule
\end{tabular}
\caption{chrF/metricX score for WMT 23 $En \rightarrow De$ with fine-grain CAT. Split ParaCrawl corpus by URL gives us more information about its data quality. We apply statistical significant test and mark the top-1 chrF score in \textbf{bold} text if p-value is smaller than 5\% compare to the runner-up.}
\label{table:split_paracrawl}
\end{table*}

\subsubsection{Granularity of CAT and corpus selection}
In the above sections, we group examples by their corpus names, but we can apply more fine-grained information to tag training data. For example, during data annotation, we can label data by the annotator ID to distinguish difference between each annotator. In ParaCrawl, the training examples are logged with URL, and we can group examples that come from the same web-domain assuming examples crawled from the same web-domain share similar distribution. With that, CAT achieves a lot finer granularity. The web-domain follows a long tail distribution and we only tag the top 1000 domains and group all the remaining URL under the same `<ParaCrawl-other>` tag. The distribution is attached in \Cref{sec:paracrawl_details}.

In \Cref{table:split_paracrawl}, splitting the corpus further reveals more details, and we can leverage CAT evaluation results to understand the quality and domain matching between the evaluation set and training corpus. For example, among all paraCrawl datasets, \texttt{Baunat} seems to have the best translation quality and bitext from \texttt{5zixi} seems to be the worst. Eyeballing some of the bitext, \texttt{5zixi} dataset contains many misaligned sentences. We share randomly sampled training data in \Cref{sec:paracrawl_details} and do not cherry-pick examples.

We can also design some heuristics to automatically select high quality datasets for fine-tuning based on CAT evaluation restuls, or filter out low quality corpus in the training. For WMT23 en-zh, the highest quality corpora that matches evaluation set is news-commentary-v18 which aligns with our expectation. In real world scenarios, we may have more datasets that would be costly for corpus selection by human experts. CAT evaluation can be one of the potential approaches to efficiently and effectively achieve the same goal. We leave this part as future work.

\section{Conclusion and Future Works}
In this paper, we summarize corpus aware training (CAT) as a generic approach to utilize corpus information. We propose an easy, efficient and effective way to fine-tune a CAT pretrained model -- OCAT. We showed OCAT successfully leverages high quality, small quantity fine-tune data which boosts model prediction significantly in machine translation setup (WMT) and mitigate inference tag selection problem of CAT.

\section{Limitations}
This work only focus on models pre-trained with CAT, and is limited to scenarios where there are multiple different sources and distribution of training data. If there is no distribution difference between corpus tags, eg. all the training data is distilled from the same teacher model then there will be no impact of switching corpus tags and therefore no benefits of fine-tuning the single tag embedding.

Although OCAT is a generic approach, we limit this work to machine translation task with unidirectional enc-dec transformer base, and do not explore more languages, machine learning tasks, model architectures nor model sizes. We also find most of the OCAT gains are limited to chrF rather than metricX.

\bibliography{custom}
\bibliographystyle{acl_natbib}

\clearpage
\onecolumn
\appendix

\begin{table*}[htb!]
\centering
\def\rownumber{}
\begin{tabular}{@{\makebox[2em][l]{\rownumber\space}}llcccc}\toprule
 Model     & inference tag         & nt23 & nt22 & flores devtest & flores dev
 \gdef\rownumber{\stepcounter{magicrownumbers}\arabic{magicrownumbers}} \\\midrule
 w/o CAT   & -                     &  58.1 / 4.5  &  60.3 / 1.2  &  64.1 / 1.1  &  63.3 / 1.1  \\ 
 + full FT & -                     &  42.3 / 7.3  &  \textbf{100.0} / 0.9  &  49.0 / 2.2  &  49.9 / 2.0  \\ 
 + adapter & -                     &  58.5 / 3.7  &  65.4 / 1.0  &  64.4 / 1.0  &  63.7 / 0.9  \\ 
 + LoRA    & -                     &  54.0 / 4.2  &  77.9 / 1.0  &  62.3 / 1.0  &  61.8 / 1.0  \\ 
\midrule
 w/  CAT & <ParaCrawl-eurowings> &  59.2 / 4.1  &  60.5 / 1.2  &  64.4 / 1.0  &  63.8 / 1.0  \\ 
         & <ParaCrawl-maoyt>     &  57.4 / 4.6  &  54.6 / 1.5  &  60.1 / 1.4  &  59.2 / 1.3  \\ 
         & <ParaCrawl-other>     &  58.9 / 4.2  &  60.5 / 1.2  &  64.5 / 1.1  &  63.9 / 1.0  \\ 
         & <news-commentary-v18> &  58.8 / 4.1  &  59.9 / 1.1  &  65.3 / 1.0  &  64.4 / 1.0  \\ 
         & <rapid2016>           &  52.1 / 5.8  &  59.2 / 1.2  &  63.8 / 1.1  &  62.8 / 1.0  \\ 
         & <wikimatrix-v1>       &  48.3 / 8.4  &  56.1 / 1.6  &  62.0 / 1.3  &  61.3 / 1.3  \\ 
  \midrule
 +  OCAT &                       &  \textbf{59.9} / 3.9  &  61.6 / 1.1  &  65.3 / 1.0  &  64.5 / 1.0  \\ 
\bottomrule
\end{tabular}

\caption{chrF/metricX score for WMT 23 $En \rightarrow De$ trained w/ \& w/o CAT and follow OCAT fine-tuning with newstest22 + flores.dev. Similar to above tables, we only mark numbers in \textbf{bold} font if it is statistical significant better than the runner-up by p-value less than 5\%.}
\label{table:cat_en_de}
\end{table*}

\section{WMT 23 En-De}
\label{sec:en_de}

We apply the same exploration for WMT 23 en-de with the same setup as above WMT23 en-zh experiment. There are 13 data sources in WMT en-de (euroParl-v10, paraCrawl-v9, commonCrawl, news-commentary-v18, wikiTitles-v3, airbaltic-v1, czechtourism-v1, ecb2017, eesc2017, ema2016, rapid2016, rapid2019, wikiMatrix-v1) and we applied the above fine-grained approach to split paraCrawl corpus by top 1000 URL domains. Therefore, there are 13 + 1000 corpus tags added to the dictionary. Note that there are 278M sentences in WMT En-De paraCrawl which takes over 90\% of the whole dataset, and training data is overwhelmed by this single dataset. Even if we split paraCrawl by URL domains, due to the long tail distribution, there are 222M sentences under paraCrawl-other corpus tag. We apply a hard 10M sentence limit per corpus tag so CAT can be more effective. We report only a few inference tag chrF scores in \Cref{table:cat_en_de} and noticed that decoding most of ParaCrawl tags show nice chrF scores except a few web-domains. We suspect that there is more rigorous filtering applied to ParaCrawl En-De than En-Zh and filtered corpus shared much similar distribution.

We also compare different fine-tuning approach against OCAT and only use newstest22 as fine-tune dataset to verify the impact of domain difference (newstest23 as in-domain and flores as out-domain). Similar to WMT23 en-zh, we also notice that full-weight, adapter and LoRA fine-tuning require careful hyperparameter setting to avoid overfitting, and OCAT is more robust to hyperparameters and resilient to overfitting. In \Cref{table:cat_en_de} row 1-4 and last row, OCAT is significantly better than all the other fine-tune techniques for both in-domain (newstest23) and out-domain (flores).

\begin{table}[h]
\centering
\begin{tabular}{|c c|} 
 \hline
domains & \# of sents \\
\hline
others & 9036858 \\
blogspot & 641582 \\
wikipedia & 540739 \\
tripadvisor & 334227 \\
alldatasheetcn & 234290 \\
freelancer & 168399 \\
ferryto & 142338 \\
booking & 141629 \\
ibooked & 130292 \\
agoda & 127487 \\
bashny & 122339 \\
taiwantrade & 99145 \\
english-video & 87683 \\
ebay & 77420 \\
jukuu & 73531 \\
rlhymersjr & 72444 \\
wikifun & 61232 \\
fnetravel & 60607 \\
bbintl & 59736 \\
wikihow & 52530 \\
agriaffaires & 48053 \\
 \hline
\end{tabular}
\caption{ParaCrawl En-Zh web domain distributions sorted by sentence count. We only pick the top 1000 domains and merge remaining domains to `others`, which is the largest source due to long tail distribution.}
\label{table:en_zh_domains}
\end{table}

\section{En-Zh ParaCrawl details}
\label{sec:paracrawl_details}
We share the top 20 out of 1000 web domains splitting En-Zh ParaCrawl by URLs in \Cref{table:en_zh_domains}, and randomly sample a few examples from selected domains (baunat, php, 5zixi) to analyze data quality (no cherry-pick applied) in \Cref{fig:sample}.

\begin{figure*}[htb]
    \centering
\scriptsize \begin{alltt}
=== baunat ===
S-1   Invest in diamonds? Diamond price
T-1   \zh{投资钻石? 钻石价格}

S-2   I can defintelly recommend Baunat to everyone, the service is beyond excellent, the diamond earrings are beautiful and the best
part is  that the prices are 2 - 3 times cheaper than you will find at the traditional (physical) jewelleries.
T-2   \zh{我 可以绝对推荐Baunat给大家,服务比出色更出色,钻石耳钉不用说非常美,最好的是,你会发现,他们的价格比在传统珠宝店要便宜2到3倍。 更
多消息 Jim}

S-3   2.90 carat diamond earrings in yellow gold
T-3   \zh{经典系列 2.90克拉白金钻石耳环}

S-4   Autumn wedding: adding colour through gemstone set gold jewellery
T-4   \zh{秋季婚礼:以镶嵌宝石的珠宝增添色彩}

S-5   Cristian Ring purchase Very good shopping experience from my visit to the showroom in Paris to delivery.
T-5   \zh{我从参观巴黎展示厅到最终收到产品,整个过程的购物体验非常不错。}


=== php ===
S-1   // Examples:
T-1   / / test function:

S-2   Important details of Sarawilliams
T-2   ammber\zh{重要详情}

S-3   All messages express the views of the author, and neither the owners of bulk-online Forums, nor vBulletin Solutions, Inc. 
(developers of vBulletin) will be held responsible for the content of any message.
T-3   \zh{所有消息都表达了作者的意见和业主的美 国恶霸显示狗 & 论坛 (中文 ) , 也不 vBulletin 解决方案,公司 (vBulletin 的开发人员) 将对
任何消息的内容负责。}

S-4   PHP: usort - Manual
T-4   PHP: \zh{数组} - Manual

S-5   This ensures that an application will be able to use the same data access paradigm regardless of the capabilities of the database.
T-5   \zh{这样可以确保不管数据库是否具有这样的功能,都可以确保应用程序可以用相同的数据访问模式。}


=== 5zixi ===
S-1   He smiled, deciding he didn’t have to worry.
T-1   \zh{到了自己儿子这里,淑嘉却免不得要担心了。}

S-2   Such 'leaders' become the sole arbiters of all truth.
T-2   \zh{这句话自此成为人本主义的口号。}

S-3   But why do you ask, Leo?
T-3   \zh{但你们为什么会这么赶时间呢 ? "}

S-4   Why is the Lerner e-mails so important?
T-4   \zh{所谓的自由究竟在哪里 ? "}

S-5   What is wrong with you anyway?
T-5   \zh{这是什么成例?}

\end{alltt}
    \caption{Random sample bitext from 3 different web domains in paraCrawl En-Zh.}
    \label{fig:sample}
\end{figure*}

\end{document}